\documentclass[10pt, conference, compsocconf]{IEEEtran}
\ifCLASSINFOpdf
\else
\fi
\usepackage{graphicx}
\usepackage{amsmath}
\usepackage{booktabs}

\begin{document}
\topmargin=0mm
%

\title{Unravelling Small Sample Size Problems in the Deep Learning World}



\author{
\IEEEauthorblockN{Rohit Keshari$^+$, Soumyadeep Ghosh$^+$, Saheb Chhabra$^+$, Mayank Vatsa$^*$, Richa Singh$^*$}
\IEEEauthorblockA{+ IIIT Delhi, India, * IIT Jodhpur, India\\
\{rohitk, soumyadeepg, sahebc\}@iiitd.ac.in, \{mvatsa, richa\}@iitj.ac.in}
}


%


\maketitle

\begin{abstract}
The growth and success of deep learning approaches can be attributed to two major factors: availability of hardware resources and availability of large number of training samples. For problems with large training databases, deep learning models have achieved superlative performances. However, there are a lot of \textit{small sample size or $S^3$} problems for which it is not feasible to collect large training databases. It has been observed that deep learning models do not generalize well on $S^3$ problems and specialized solutions are required. 
In this paper, we first present a review of deep learning algorithms for small sample size problems in which the algorithms are segregated according to the space in which they operate, i.e. input space, model space, and feature space. 
Secondly, we present Dynamic Attention Pooling approach which focuses on extracting global information from the most discriminative sub-part of the feature map. The performance of the proposed dynamic attention pooling is analyzed with state-of-the-art ResNet model on relatively small publicly available datasets such as SVHN, C10, C100, and TinyImageNet.  

\end{abstract}


\section{Introduction}

\begin{figure}[!t]
	\centering
	\includegraphics[width=0.47\textwidth]{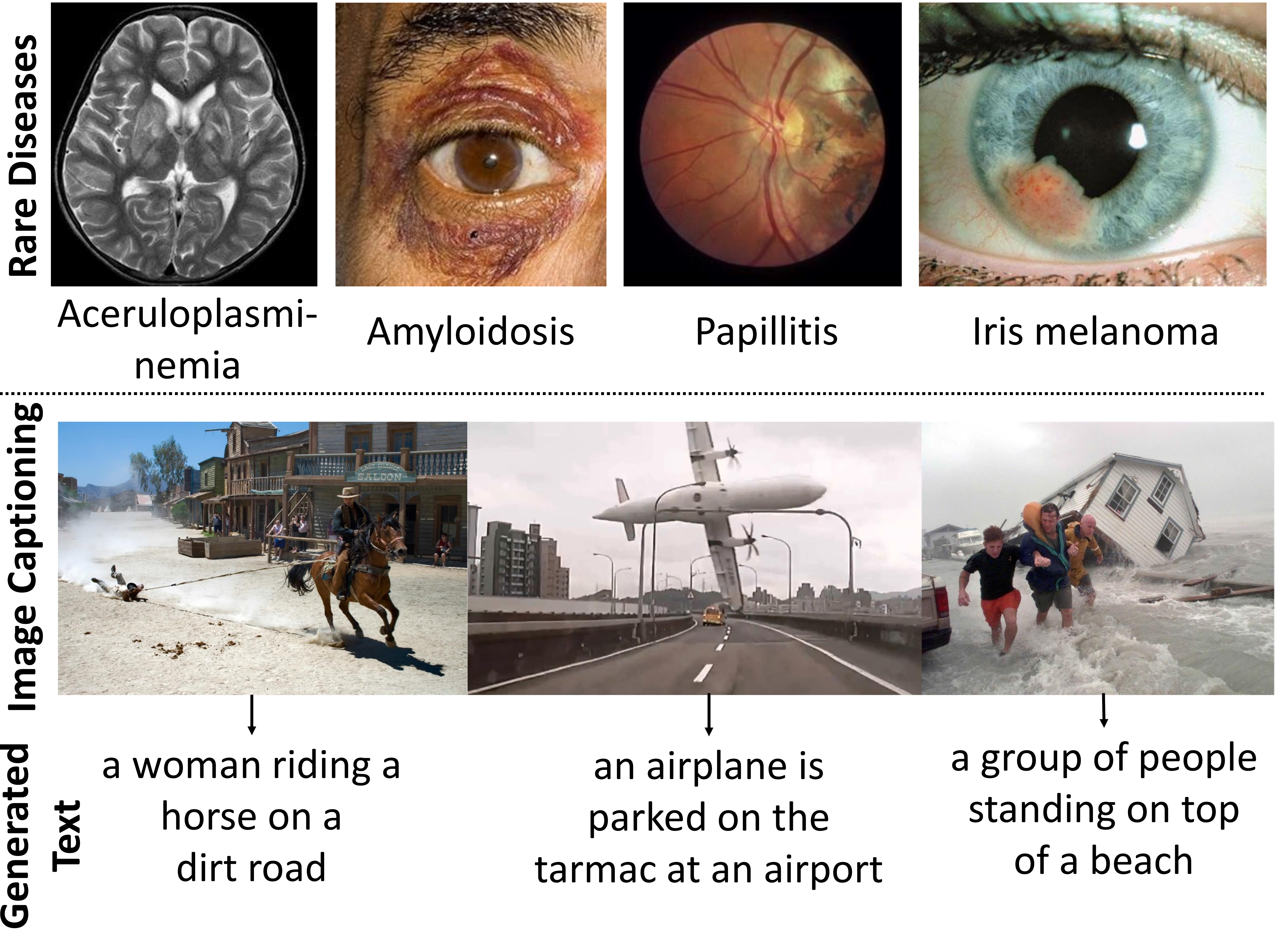}
	\caption[]{Illustrates samples of rare diseases and caption generation from scenes. (\textbf{Top}). Aceruloplasminemia\footnotemark[1], Amyloidosis\footnotemark[2], Papillitis~\cite{eggenberger2014neuro}, Iris melanoma\footnotemark[3]. (\textbf{Bottom}) The captions have been generated by a deep neural network code\footnotemark[4]. From the images, it can be observed that intuition and concept are missing in the generated captions. \textit{Image credit:} ~\cite{lake2017building}.}
		\label{fig:motivation}
	
\end{figure}

One of the first requirements in building a machine learning system is adequate high quality training data. As the research community has progressed, this requirement has become an integral part, and the currently popular deep learning models require a very large number of samples for training ~\cite{he2016deep,huang2017densely,majumdar2016face}. However, collecting large number of samples and annotating them is not always feasible. For instance, Figure~\ref{fig:motivation} (\textbf{top}) shows images of rare eye diseases. Having multiple samples of rare diseases is challenging because a limited number of case studies are available \cite{thomas2020dealing}. In another example, Figure~\ref{fig:motivation} (\textbf{bottom}) shows the captions generated from the deep learning model. However, the generated text does not match with the descriptions present in the scenes due to the limited training samples of situations present in the scenes. 

The scientific problem is classified as small sample size ($S^3$) or small sample learning (SSL) problem when the available training data is not sufficient to learn the high dimensional feature and perform classification. In such cases, scarcity of data leads to overfitting and inaccurate classification outputs. It is an important research problem in the deep learning world and is rapidly gaining attention. Several solutions for small sample learning (SSL) have been proposed by the researchers and these solutions are often based on different questions, for instance, what are the database characteristics and what is the deep learning pipeline being used for classification. 

	
\begin{figure*}[!t]
		\centering
		\includegraphics[scale = 0.72]{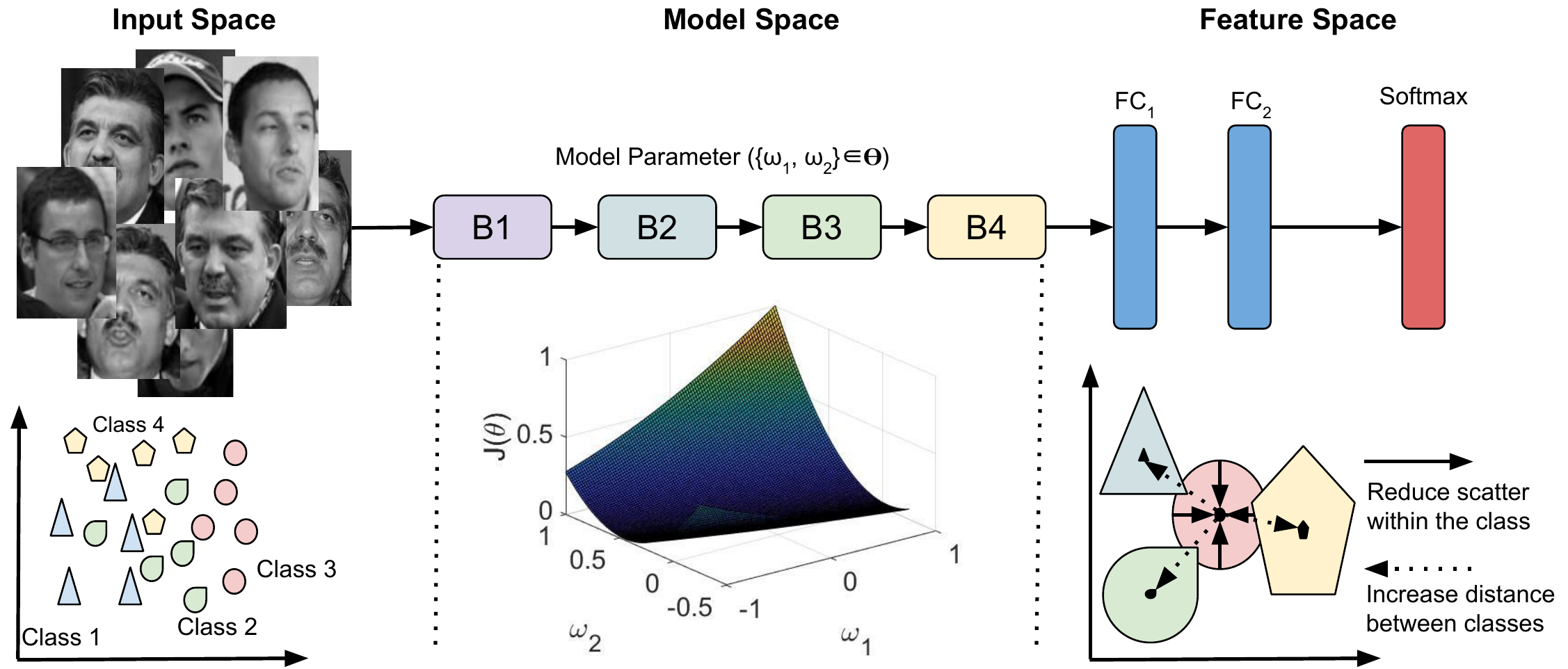}
		\caption{\small{Illustration of input space, model space, and feature space for the deep learning model. In case of a CNN model, B1 to B4 are basic blocks which contain operations such as $convolution$, $batchnorm$, and $ReLu$. $FC_1$ and $FC_2$ are fully connected layers. The last layer is a $softmax$ layer used for classification.}}	
		\label{fig:space} 	
\end{figure*}

Let us analyze the SSL problem using the general formulation of deep learning classifiers. Mathematically, a deep learning model can be represented as:

\begin{equation}
    \mathbf{F} = \phi(\mathbf{WX} + b)
\end{equation}
where, $\phi$ is the model with weights $\mathbf{W}$ and bias $b$. This model takes $\mathbf{X}$ as input and outputs the feature $\mathbf{F}$. We first propose that the SSL algorithms can be categorized into \textit{input, model}, and \textit{feature space} depending on whether they are operating at $\mathbf{X}$, $\mathbf{W}$ and $b$, or $\mathbf{F}$, respectively (Figure \ref{fig:space}). Input space refers to the set of algorithms which increases the database by generating more samples or perturb the samples to optimize the feature space \cite{suri2018matching,bousmalis2017unsupervised,huang2019generative,chhabra2019data}. In the feature space, the algorithms operate on representations by reducing the intra-class distance and maximizing the inter-class distance to improve the classification performance \cite{hadsell2006dimensionality,schroff2015facenet,chen2017beyond}. The algorithms operating in the model space approximate the target function to map inputs to the outputs~\cite{he2016deep,huang2017densely}. For better learning and generalization of the model or target function, several regularization algorithms have also been proposed \cite{srivastava2014dropout,wan2013regularization,keshari2019guided}. The first contribution of this paper is a summary of the SSL algorithms according to the input, model, and feature space.   

The second contribution of this research is the proposed \textbf{Dynamic Attention Pooling (DAP)} to improve the generalizability of the model. We propose to apply DAP in place of global pooling at the last layer of CNN architecture. It drops the features map selectively that leads to the removal of insignificant features maps thereby improving the generalizability of the model even when the models are trained on relatively small databases. The next three sections provide the review of the SSL algorithms according to the proposed categorization of input, model and feature spaces. Section \ref{sec:dap} presents the proposed Dynamic attention pooling along with the results obtained on four databases and key observations.
\section{Optimizing Input Space for SSL Problems}
In machine (deep) learning, it is important to understand the role of input space while learning the model space and feature space. Generally, the input space grows exponentially with the increase in dimension which makes the ML problem intractable \cite{domingos2012few}. For instance, for a $100$ dimensional binary input data, there are $2^{100}$ possible inputs. A training set with trillion examples covers only $10^{-18}$ part of the input space which is a very small fraction of the input space. 
Machine learning models for $S^3$ problems can be optimized if we can address, to an extent, data variations in the input space. These data variations are broadly correspond to applications such as domain adaptation and zero shot learning. In this section, we present an overview of data augmentation or alteration approaches which address the challenges related to supervised and unsupervised domain adaptation as well as zero shot learning. Further, we discuss the concept of data fine-tuning to enhance the performance of pre-trained models.

\subsection{Data Augmentation}
For Small Sample Learning problems, the role of input space has been well explored in the literature of domain adaptation via data augmentation. The idea is to form an augmented dataset in the target domain to compensate for small samples by transforming and augmenting the data from the source domain with certain constraints. \cite{sugiyama2008direct} addressed the problem of covariate shift when the input training samples and testing samples follow different distributions. In conventional methods, the importance factor is used to quantize the covariate shift by accurately estimating the ratio between training and testing input density. However, in higher-dimensional cases, it becomes hard to estimate the data density. Therefore, the authors proposed the cross-validation based techniques to directly compute the importance factor. 
In order to measure the discrepancy between source and target domains, maximum mean discrepancy (MMD) based methods \cite{baktashmotlagh2016distribution}, \cite{long2013transfer}, \cite{pan2010domain}, \cite{tzeng2014deep} are widely used by researchers. However, these methods ignore the class weight bias across domains. To address this issue, \cite{Yan_2017_CVPR} proposed a weighted MMD model which exploits the class prior probabilities of source and target domains. 

Saenko \textit{et al.} introduced a new line of research based on the transformation between the source and target domains \cite{saenko2010adapting}. The aim is to learn a mapping between the points of the source domain and target domain in a supervised manner. Long \textit{et al.} \cite{long2014transfer} have shown that both feature matching and instance re-weighting play a key role in visual domain adaptation. Hence, they propose transfer joint matching (TJM) approach which reduces the difference across domains by matching features and instance re-weighting in a dimensionality reduction manner. The proposed approach outputs a representation which is invariant to both distribution difference and irrelevant instances.
Das and Lee \cite{das2018sample} have proposed technique for unsupervised domain adaptation which finds the correspondence between samples of source and target domains. The samples in both domains are treated as graphs and convex criteria are used to match them. Class-based regularization and the first and second-order similarities are used as a criterion between graphs of both the domains. Learning a mapping function has also been utilized to solve zero/few-shot problems~\cite{atzmon2019adaptive}. The mapping learns generalized representation from the training set to classify new concepts (novel classes) correctly from the testing set. Classifying the novel classes at the inference is termed as Zero-Shot or Generalized Zero-Shot Learning (ZSL/GZSL). Soh \textit{et al.} ~\cite{soh2020meta} have proposed zero-shot-super-resolution (ZSSR) to improve the resolution of an image by exploiting both external and internal information, where one single gradient update can provide quite considerable results. Min \textit{et al.}~\cite{min2020domain} have proposed Domain-aware Visual Bias Eliminating (DVBE) network by constructing two complementary visual representations; semantic-free and semantic-aligned. They have explored cross-attentive second-order visual statistics to compact the semantic-free representation. Recently, data generation process has been proposed to tackle such challenging scenarios~\cite{huang2019generative,schonfeld2019generalized,zhang2019adversarial}. Keshari \textit{et al.}~\cite{keshari2020ocd} proposed to generate an Over-Complete Distribution (OCD) using Conditional Variational Autoencoder (CVAE) of both seen and unseen classes. They observed that generating synthetically overlapped distribution and forcing the classifier to transform into non-overlapping distribution can improve the performance on both seen and unseen classes.

Existing work primarily focuses on extracting features which are domain invariant in an unsupervised domain adaptation. Bousmalis \textit{et al.} \cite{bousmalis2017unsupervised} have proposed a novel technique which learns the transformation in pixel space between source and target domains in an unsupervised manner. They propose a Generative Adversarial Network (GAN) based architecture which learns to map the images from the source domain to target domain such that the images are drawn from the target domain. Their method, termed as Pixel level Domain Adaptation (PixelDA), does not require one-to-one correspondence between samples of source and target domain. Taigman propose a Domain Transfer Network (DTN) \cite{taigman2016unsupervised} which maps a sample from the source domain to an analogous sample in target domain such that the output of a function which takes images from either domain remains same. This technique is performed in an unsupervised manner. Tzeng \textit{et al.} \cite{tzeng2017adversarial} show that the generative adversarial networks are not good in discriminative tasks and are limited to smaller domain shifts. To address this problem, they proposed a generalized framework for adversarial adaptation which uses discriminative modeling, untied weight sharing, and a GAN loss and termed it as Adversarial Discriminative Domain Adaptation (ADDA). Murez \textit{et al.} \cite{murez2018image} have used the unpaired image-to-image translation framework and proposed a method to constrain the extracted features from the encoder network such that they are able to reconstruct the images in both source and target domains. 
Recently, Zhang \textit{et al.} \cite{zhang2020deep} proposed deep adversarial data augmentation (DADA) technique to address the problem of extremely low data regimes. DADA enforces both real and augmented samples to contribute in finding the decision boundaries.

\subsection{Data Fine-tuning} 
Pre-trained models are widely used for ML tasks ranging from face recognition to object classification and segmentation. However, these pre-trained networks generally do not yield good performance if the target database is different from the source (pre-training) database. For example, a Convolutional Neural Network (CNN) pre-trained on VGGFace2 database \cite{cao2018vggface2} may not yield good results when tested with ImageNet database \cite{deng2009imagenet}. Two methods are widely used for fine-tuning pre-trained CNN models: (i) freezing the convolutional layers and training the dense layers added after the convolutional layers (ii) re-training few convolutional layers along with the dense layers by updating weights while leaving the other convolutional layers frozen. However, the number of trainable parameters in these methods are large, especially for deeper models such as ResNet-150 \cite{he2016deep} and DenseNet-201 \cite{huang2017densely}. Chhabra \textit{et al.} \cite{chhabra2019data} recently have proposed data fine-tuning (DFT) which leverages the input space to enhance the performance of the pre-trained CNN models. In this technique, input data (target database) is ``adjusted'' corresponding to the pre-trained model's unseen decision boundary. 
To illustrate this concept, let $\phi$ be the pre-trained model with weight $\mathbf{W}$ and bias $b$. DFT can be represented as,
\begin{eqnarray}
\phi(\mathbf{WX} + b) \xrightarrow[\text{}]{\text{DFT}} \phi(\mathbf{WZ} + b)
\end{eqnarray}
where, $\mathbf{Z}$ represents the updated dataset. 
To adjust the input data, a uniform perturbation is learned corresponding to each dataset. Comparing it to model fine-tuning (MFT), MFT involves updating $\mathbf{W}$ and $b$ whereas in DFT, input data $\mathbf{X}$ is updated. 
It is important to note that the number of trainable parameters in DFT is same as the size of the input image, which is significantly less compared to the number of trainable parameters in deep learning models.

\section{Optimizing Model Space for SSL Problems}

Large training samples play an important role in successfully training deep architectures, which has millions of learning parameters~\cite{he2016deep,huang2017densely,szegedy2015going}. The parameters of the deep neural network can be represented as $\theta$, $x$ is input, and $y$ is ground truth for $x$. Mathematically, learning of a network can be represented as:


\begin{equation}
	J(\theta)=\underbrace{\frac{1}{m}\sum_{i=1}^{m} Cost(h_{\theta}(x^{(i)}),y^{(i)})}_\text{Model Specific}+\underbrace{R(\theta)}_\text{Domain Specific}
	\label{eq:mleq}
\end{equation}

\noindent where, $m$ is the number of training samples, $h_{\theta}$ is the hypothesis $h_\theta(x)\in [0,1]$, and $R$ is the regularizer constrained on $\theta$ while learning. 
To avoid overfitting due to SSL, researchers have proposed rectification of the pre-trained model and strong regularization (to reduce the dependency on Large Sample Learning (LSL)).

\subsection{Adapting Pre-trained Models}
In the literature, adaptation of the model has been achieved by fine-tuning, distillation, and model adaptation of the pre-trained model. Generally, large datasets can be utilized for pre-training the models. 
In \textit{fine-tuning}, models pre-trained for a similar task can be re-trained on small sample data (target dataset)~\cite{krizhevsky2012imagenet,hinton2006reducing,yosinski2014transferable}. Hinton and Salakhutdinov~\cite{hinton2006reducing} suggested to train the Stacked Auto-Encoder (SAE) in-place of training the whole model at a time. Therefore, overfitting and vanishing gradient problem can be minimized. Inspired by SAE,~\cite{bharadwaj2016domain} have utilized Stacked Denoising Auto-Encoder (SDAE) for newborn face recognition which has a small number of training samples. Stack-wise learning can also be utilized in the dictionary learning paradigm. Tariyal \textit{et al.} \cite{tariyal2016deep} have shown that popular deep learning models can be designed with the help of dictionary, hence, can be used in SSL. In CNN~\cite{yosinski2014transferable}, the concept of fine-tuning of few layers works because of learning of common feature extractor at the initial layer of CNN model, irrespective of databases. The initial layer of the CNN model learns Gaussian type filters to extract the edge and blob based information. After the initial layers, subsequent layers learn complex feature extractor which can provide the abstract view of the object~\cite{ren2015faster,chu2016best}. 
    
Recently, researchers have proposed several techniques which outperform the performance of conventional MFT. This includes progressive network~\cite{rusu2016progressive}, block-module network~\cite{terekhov2015knowledge}, utilizing intermediate information of the CNN blocks~\cite{lakra2018segdensenet}, class-based penalty at each convolutional layer~\cite{siddiqui2018face}, and collaborative learning~\cite{goellc}. Keshari \textit{et al.}~\cite{keshari2018learning} have observed that the structure of the CNN filters can be learned separately. Hence, the learned filters are used to initialize the targeted model. After initializing the learned filters in CNN framework, the parameters of the filters are frozen and only strength of a filter has been trained. The experiments support our hypothesis that learning only the strength of filters can reduce the total learning parameters, hence the performance of the model on small sample data can improve. These mentioned networks are either trained on multiple tasks to have a more generalized network or divide the whole learning parameters into modules and trained the network in a modular fashion. Recently, to address $S^3$ problem, \cite{d2020structural} have proposed VC-dimension based network structure optimization for CNNs.

\textit{Distillation} of the model is similar to knowledge transfer method~\cite{hinton2015distilling}. However, distillation has Teacher-Student Network (TSN), where the student network is trained with feedback from the teacher network. The feedback has been introduced by reducing the cross-entropy between the softened output of teacher network to the output of the student network. Romero \textit{et al.}~\cite{romero2014fitnets} have proposed to utilize not only the softened output but also the intermediate representation of the teacher network. Similarly, Yim \textit{et al.}~\cite{yim2017gift} have proposed to utilize the flow between layers of the network by computing the Gram matrix of both consecutive layers. Radosavovic \textit{et al.} ~\cite{radosavovic2018data} have proposed Omni-supervision for distillation of the network. In this method, unlabeled data has been used to generate new training data while predicting multiple transform data from a pre-trained model. Luo \textit{et al.}~\cite{luo2018graph} have proposed graph-based distillation of the model for partially observed modalities.

In \textit{model adaptation}, unlike the data transportation, decision boundary of the model is adapted on the target dataset which has less number of observation. To do the model adaptation in SVM, domain transfer~\cite{duan2009domain}, cross-domain~\cite{jiang2008cross}, residual transfer~\cite{long2016unsupervised}, and adaptive multiple kernel learning~\cite{duan2012exploiting} have been proposed. Tzeng \textit{et al.}~\cite{tzeng2015simultaneous} have utilized two losses (softmax cross-entropy loss, and domain confusion loss) to train the network on target dataset. Moreover, in place of confusion loss, Long \textit{et al.}~\cite{long2015learning} have used Maximum Mean Discrepancy (MMD) loss on fully connected layers' of AlexNet~\cite{krizhevsky2012imagenet}. Sener \textit{et al.}~\cite{sener2016learning} have proposed to optimize the target label inference along with features in the deep network such that domain transfer on small sample data can be done.

\subsection{Reduce the Dependency of Large Sample Learning}
In general, learning mechanism of machine learning models follow Equation~\ref{eq:mleq} which can be seen as a combination of 1) model specific learning and 2) domain specific regularization. The multi-attention framework proposed by Huynh and Elhamifar~\cite{huynh2020shared} can be considered as model-specific learning. In this work, they have introduced a shared multi-attention framework for multi-label zero-shot learning. They have demonstrated that having attention mechanism for recognizing multiple seen/unseen labels is a complex task. Hence, instead of generating attention for novel classes, they have let the novel classes select from a set of shared attentions. Generally, regularizers are considered as domain-specific knowledge used to improve the training of the network. In this school of thought, Tenenbaum \textit{et al.}~\cite{tenenbaum2011grow} have suggested that strong prior makes a difference of making inferences beyond the data availability. This prior can be side information~\cite{vapnik2009new}, domain knowledge~\cite{pan2010survey}, and common sense~\cite{davis2015commonsense}. In the literature, dropout~\cite{srivastava2014dropout}, drop-connect~\cite{wan2013regularization}, batchnorm~\cite{ioffe2015batch}, class based sparsity~\cite{sankaran2017class, sankaranIVC, gauravFusion, aparnatifs}, and guided-dropout~\cite{keshari2019guided} have been proposed to reduce the dependency on LSL by reducing  overfitting. Keshari \textit{et al.} \cite{keshari2019guided} have recently proposed strength-based guidance of dropping nodes in the training phase. They have observed that most of the nodes are in-active and therefore, dropping them in a guided fashion can improve the over-all performance of the model.



\section{Optimizing Feature Space for SSL Problems}
In the last decade, advancements in deep learning algorithms have enabled the realization of several real world applications ranging from face and gesture recognition to autonomous driving vehicles and drones. Face recognition from surveillance cameras is important to ensure public safety and avoid instances of terrorist attacks and intrusion. One of the primary reasons for the advancement of such applications is the advent of novel and effective loss functions that is focused primarily on the feature space. These loss functions update the parameters of the model such that it produces feature rich representation in the embedding space of the model. The most prolific of these are loss functions formulated using Deep Metric Learning (DML) which allow us to train discriminative classifiers even on databases with insufficient training samples. This section highlights DML as an effective technique for training discriminative models for $S^3$ problems.

\subsection{Deep Metric Learning (DML)}
Conventional deep learning models like Convolutional Neural Networks (CNNs) and Recurrent Neural Networks (RNNs) are trained with data samples along with their corresponding labels, so that they can correctly predict the class/label of an input sample during testing. However, deep metric learning algorithms train a model with the objective of distinguishing between a pair of data samples whether they belong to the same class/category or not. During training, a vanilla deep metric learning loss function would update the weights of the model so that it produces embeddings/features (unlike class labels in conventional deep learning models) of data samples that belong to the same class close to each other, and that of different classes away from each other in the output embedding space of the model. In order to train such a model, we need large quantities of data samples during training. Since a discriminative model is being trained, it may be tested/evaluated on classes that are not encountered by the model during training. This flexibility makes deep metric learning models a popular choice for building real world recognition systems.   


The most seminal work in deep metric learning was by Hadsel \textit{et al.}~\cite{hadsell2006dimensionality} where the contrastive loss was proposed. It optimized a Siamese network for matching a pair of images, by the same optimization goal as illustrated above. 
Thereafter, several research works~\cite{sun2014deep,zheng2020deep, ma2020discriminative} have utilized this optimization technique using a deep-CNN network as the backbone model, before a new loss function, known as the triplet loss was proposed by ~\cite{schroff2015facenet}. This was extended in 2017, by a new loss function known as the quadruplet loss~\cite{chen2017beyond}. An N-pair loss metric~\cite{sohn2016improved} is also proposed which uses an N-tuple as a training data sample. Further, different variants of these loss functions are proposed for SSL scenarios.

\subsection{DML for Small Sample Learning}
In order to address one shot and few shot learning scenarios, several deep metric learning algorithms have been proposed for small sample learning. Vinyals \textit{et al.}~\cite{vinyals2016matching} have proposed an algorithm which is used to train the model in episodic cycles. In each training cycle or episode few training examples are selected to learn embeddings for predicting the class of these samples. The purpose of this episodic training strategy is to mimic the real environment where only a few samples would be available. Several other one-shot and few-shot based DML approaches~\cite{snell2017prototypical,sung2018learning,altae2017low} have also been proposed to address small sample training scenarios. Recently a density aware deep metric learning algorithm is proposed~\cite{ghosh2019cvpr} where results are shown on a surveillance face dataset (SCface~\cite{grgic2011scface}) which has a very small number of training samples. This algorithm has a mechanism of avoiding outliers and noisy training data, which can hinder the learning process, especially in SSL scenarios.

\subsection{Sample Mining in DML}
Train deep network using loss functions such as the triplet loss requires preparing triplets (or 3-tuples) using the data available for training. Given $N$ training classes and $K$ samples for each class, the total number of triplets that can be prepared for training is upper bounded by $N(N-1)K^2(K-1)$. Therefore, the number of training samples (each triplet is treated as a training sample) increases from $O(N.K)$ (available for conventional deep learning algorithms) to $O(N^2.K^3)$ which is a very large sample space. For quadruplets or N-pair loss functions, this space would be even larger. This increased sample space is extremely useful for learning a model with a DML algorithm on a database that has a small number of data samples. It also makes DML algorithms a natural choice where the amount of training data is insufficient to learn a conventional classifier.

On large databases this enormous input sample space may hinder efficient learning due to several reasons. One of the reasons is that, after several epochs of training, the model would have learnt to solve most of the data samples, each of which is a triplet/quadruplet. Thus, fewer samples would be useful for the model to continue learning and make significant weight updates. During this phase it is required to provide only useful triplets/quadruplets, in other words mine those triplets/quadruplets which are hard (which is still not correctly classified by the model) in order to continue learning the model. This technique known as hard mining, has been extensively explored  in the last few years for DML methods \cite{shrivastava2016training,yuan2017hard,suh2019stochastic,harwood2017smart,oh2016deep}.

\subsection{Adversarial Deep Metric Learning}
Recently, a new way of applying DML algorithms to small databases has been through deep adversarial metric learning. This is proposed by Duan \textit{et al.}~\cite{duan2018deep}, where synthetic data samples are generated. As illustrated above, hard-mining approaches mine hard triplets/quadruplets from the existing pool of training data. However, at some point of time the pool of training data would be exhausted, especially for small databases. This technique generates new synthetic samples from the existing data samples using a generator by an adversarial loss function. New triplets/quadruplets can then be generated from these synthetic examples, most of which are generated as hard samples. Keshari \textit{et al.}~\cite{keshari2020ocd} have utilized metric learning in ZSL/GZSL setting as well. On the generated over-complete distribution, they have proposed a framework which utilizes Online Batch Triplet Loss (OBTL) and Center Loss (CL) to enforce the separability between classes and reduce the class scatter. Another technique with a similar objective is Energy Confused Adversarial Metric Learning~\cite{duan2018deep}, where synthetic samples are generated using an energy confusion regularization term in order to confuse the DML model. This energy confusion term is trained together with the conventional metric objective in an adversarial manner. Recently an adaptive margin based approach \cite{wang2020adaptive} have also been proposed for the same. 

\section{Dynamic Attention Pooling}
\label{sec:dap}

As mentioned above, feature space approaches can be used to solve $S^3$ problems. In this section, we present the formulation of the proposed dynamic attention pooling (DAP) which is a feature space approach. We also present the implementation details and the results on multiple databases. 

\subsection{DAP Formulation}
State-of-the-art CNN architectures generally have global average pooling operation which is applied at the last layer of the network. Broadly, to classify an image using CNN, the features ($f_{ext}$) extracted by model is followed by a classification module. 
Each layer ($l$) produces a three dimensional feature map ($X^l_{i,j,k}$) represented as:

\begin{equation}
\label{eq:lmap}
\textbf{y}^l=f_{ext}^l(\textbf{X})
\end{equation}
The final score for an input image is computed by a classification function defined as $score=f_{classifier}(\textbf{y}^l)$. The function $f_{ext}$ has multiple layers of convolutional filters and non-linearity function such as $ReLu$ while $f_{classifier}$ has multiple fully-connected layers or a global pooling layer followed by the  softmax layer.

\begin{figure}[!t]
	\centering	
	\includegraphics[scale = 0.5]{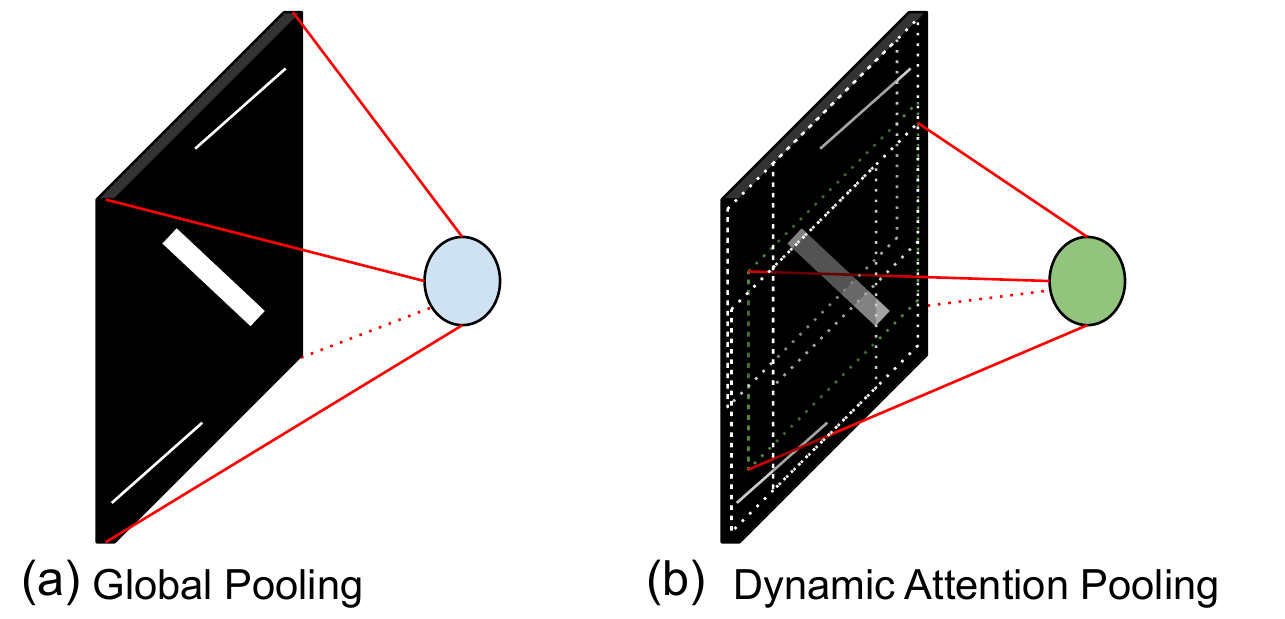}
	\caption{(a) Global average pooling operation map all the features into one value. If the feature map has sparse values and noise then averaging the whole feature map might give misleading representation. (b) Proposed DAP operation works on a small window which traverses on the feature map and seeks maximum attention to classify an object.}
	\label{fig:moti}
\end{figure}

As shown in Figure~\ref{fig:moti}, since global average pooling performs averaging operation on the complete feature-map, it might reduce the intensity of the features. To improve this, we propose dynamic attention pooling which dynamically reduces the loss computed by softmax followed by the cross-entropy loss. DAP is applied at the last layer of a CNN model for better generalizability over the unseen test data. From equation~\ref{eq:lmap}, $\textbf{y}^l$ is the feature-map at the $l^{th}$ layer. In place of global pooling, local average pooling of window $[pw, pw]$ has been applied on $\textbf{y}^l$ with $s$ stride.

\begin{figure}[!t]
 	\centering
 	\includegraphics[width=.5\textwidth]{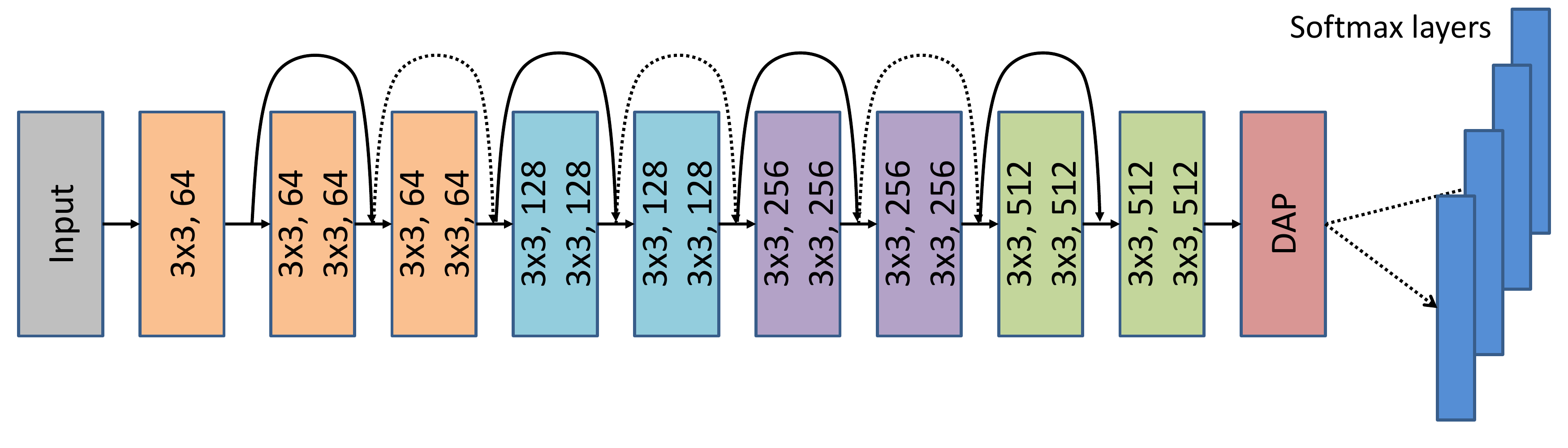}
 	\caption{Representation of ResNet18~\cite{he2016deep} with dynamic attention pooling as a last layer. Pooling on small window assign to a separate softmax layer.}
 	\label{fig:resNet18p} 	
 \end{figure}
 
\begin{equation}
\label{eq:avg_pol}
\textbf{y}^{l+1}=\frac{1}{pw\times pw}\sum_1^{pw \times pw}\left (\sum_{1}^{s:pw}\sum_{1}^{s:pw}\textbf{y}^l(pw, pw)\right )
\end{equation}
where, $\textbf{y}^{l+1}$ is the output of average pooling on a small window. Let the pooled feature-map $\textbf{y}^{l+1}$ is of resolution $n_1\times n_2$ and it is divided into $n(=n_1\times n_2)$ features vectors $\{f_1,f_2,...,f_n\}$ such that $\forall$ $f_i\in \textbf{y}^{l+1}$, where $i=1,2,\cdots,n$. Each individual feature vector $f_i$ is separately fed into the softmax layer which learns the probability $P_i$ of class $c$. 

\begin{equation}
\label{eq:softmax}
P_i(y=c|f_i)=\frac{e^{f_iW_{i,c}}}{\sum_k e^{f_iW_{i,k}}}
\end{equation}

\noindent where, $W_{i,k}$ is the weight matrix associated with the $i^{th}$ softmax layer. Cross-entropy loss of the $i^{th}$ softmax layer can be written as:

\begin{equation}
\label{eq:loss}
loss_i=\sum_k -log(\lambda_iP_i(y=k|f_i)) \quad where, i=1, 2,\cdots, n
\end{equation}
Here, $\lambda_i$ is the weight associated with the $i^{th}$ softmax layer. Based on the contributions of $f_i$, $\lambda_i$ learns the weight for the softmax layer while performing back-propagation. Note that the feature $f_i$ is computed on sub-blocks of an image and loss is optimized dynamically for the local region of an image. Mathematically, it can be expressed as:

\begin{equation}
\label{eq:loss_d}
loss=max(loss_1, loss_2,\cdots, loss_n)
\end{equation}
This dynamic routing between the loss is inspired by the seminal work of Sabour~\textit{et al.}~\cite{sabour2017dynamic} where the route is selected if and only if the desired class is present. Similarly, the obtained loss in equation~\ref{eq:loss_d} is back-propagated in the network and optimized through stochastic gradient descent (SGD)~\cite{lecun1988theoretical}. 
Finally, the classification score of an image is computed as the weighted fusion of softmax probabilities.

\begin{equation}
\label{eq:score}
score=\sum_{i=1}^n \lambda_iP_i  
\end{equation}

The weight parameter $\lambda_i$ is updated with $\Delta \lambda_i$ ($=\frac{\partial loss_i}{\partial \lambda_i}$) values and computed using partial derivation of the $loss$ with respect to $\lambda_i$.

\subsection{Implementation Details}
Experiments are performed on a workstation with two 1080Ti GPUs under PyTorch~\cite{paszke2017automatic}. The program is distributed on both the GPUs. The value of hyper-parameters such as epoch, learning rate, batch size is kept as 500, $[10^{-1},...,10^{-5}]$, and 64 respectively for all the experiments. Learning rate is started with $10^{-1}$ and reduced by a factor of $10$ at every 100 epoch. For C10/C100/SVHN, $pw$ is set as 3 and stride as $2$. For Tiny ImageNet, $pw$ is $6$ and stride is $2$. The best results have been obtained with four classifiers in the DAP formulation. Therefore, depending on the database sample resolution, pooling window $pw$ and stride are adjusted.

\subsection{Results and Analysis}
\begin{table}[]
\centering
\caption{Publicly available databases and their protocols in terms of number of classes, number of samples in training and testing sets.}
\label{tab:proto}
\begin{tabular}{|c|c|c|c|c|}
\hline
\textbf{Split}   & \textbf{C10} & \textbf{C100} & \textbf{SVHN} & \textbf{TinyImageNet} \\ \hline
\textbf{Train}   & 50k          & 50k           & 73,257        & 100k                  \\ \hline
\textbf{Test}    & 10k          & 10k           & 26,032         & 10k                   \\ \hline
\textbf{Classes} & 10           & 100           & 10            & 200                   \\ \hline
\end{tabular}
\end{table}
\begin{table}[]
\scriptsize
\centering
\caption{Test accuracies of the proposed Dynamic Attention Pooling  with ResNet18 model~\cite{he2016deep} and comparison with Gloabal Average Pooling (GAP), Gloabal Max Pooling (GMP), and combination of both pooling methods.}
\label{tb:cnn_results}
\begin{tabular}{|l|c|c|c|c|}
\hline
\textbf{Algorithm}                                                                                             & \textbf{SVHN}  & \textbf{C10} & \textbf{C100} & \textbf{TinyImageNet} \\ \hline
ResNet18 + GAP                                                                                              & 96.42          & 93.78            & 77.01             & 61.96                 \\ \hline
ResNet18 + GMP                                                                                              & 95.31          & 91.92            & 75.23             & 59.72                 \\ \hline
ResNet18 + (GMP+GAP)~\cite{pmlr-v51-lee16a}    & {96.52}          &     92.28        &   77.11           &   62.07               \\ \hline
\textbf{ResNet18 + DAP}                                                                                          & \textbf{96.52}          & \textbf{95.53}            & \textbf{77.58}             & \textbf{63.94}                 \\ \hline
\end{tabular}
\end{table}

The performance of the proposed dynamic attention pooling is evaluated on four databases, C10~\cite{krizhevsky2009learning}, C100~\cite{krizhevsky2009learning}, SVHN~\cite{netzer2011reading}, and Tiny ImageNet~\cite{timagenet} and the protocols of the databases are mentioned in the Table~\ref{tab:proto}. 
As shown in Figure~\ref{fig:resNet18p}, the ResNet18 architecture is modified at the last layer where maximum loss is propagated through a small pooling window. Table~\ref{tb:cnn_results} contains results of the proposed (DAP) and the widely used Global Average Pooling (GAP), Global Max Pooling (GMP) and the combination of both pooling methods. It can be observed that DAP with ResNet18 architecture performs better than the other three methods. On the SVHN dataset, the performance is the same as the SOTA pooling method. On C10, C100, and TinyImageNet, the improvements are around 1.75\%, 0.47\%, and 1.87\%, respectively. These results show that providing proper pooling can improve the generalization of a deep model. Specifically, in CNN, dynamic attention pooling in place of global pooling of the feature map selectively leads to the removal of insignificant feature maps. 

\section{The Way Forward}

It has been very aptly said that for every problem with a million training data points, there are a hundred problems with just a thousand training data points. Therefore, to expand the usability of machine (deep) learning algorithms, it is important to design solutions that work with fewer training samples. This paper summarizes the research efforts for the $S^3$ or \textit{SSL} problems while categorizing them into input, feature or model spaces. We have also proposed a dynamic attention pooling (DAP) in place of global pooling and supported our assertion that dropping the feature maps selectively leads to improved performance. Finally, we believe that while a lot of progress has been made in this direction; it is still a growing area with a lot of applications in socially relevant domains. Therefore, continuous efforts are required to disentangle small sample size problems in the deep learning world.

\texttt{}


\section*{Acknowledgment}
R. Keshari is partly supported by the Visvesvaraya Ph.D. Scheme of MeitY, Govt. of India. S. Ghosh is partly supported by the TCS PhD Fellowship. Research by M. Vatsa and R. Singh are partially supported by MeitY, Govt. of India. M. Vatsa is also supported through the Swarnajayanti Fellowship by the Govt. of India.

\footnotesize
\bibliographystyle{IEEEtran}
\bibliography{ijcai20}

\end{document}